\documentclass[11pt,a4paper]{article}
\usepackage[hyperref]{acl2020}
\usepackage{times}
\usepackage{latexsym}

\usepackage{microtype}

\usepackage{graphicx}
\usepackage{graphics}
\usepackage{tabularx}
\usepackage{url}
\usepackage{enumitem}
\usepackage{hyperref}

\usepackage{amssymb}% http://ctan.org/pkg/amssymb
\usepackage{pifont}% http://ctan.org/pkg/pifont
\newcommand{\cmark}{\ding{51}}%
\aclfinalcopy % Uncomment this line for the final submission

\title{Evaluating Neural Morphological Taggers for Sanskrit}

\author{Ashim Gupta\textsuperscript{1}, Amrith Krishna\textsuperscript{2}, Pawan Goyal\textsuperscript{3}, Oliver Hellwig\textsuperscript{4} \\
  \textsuperscript{1}School of Computing, University of Utah\\
  \textsuperscript{2}ITU Copenhangen, \textsuperscript{3}Dept. of Computer Science and Engineering, IIT Kharagpur\\
  \textsuperscript{4}University of Zürich, IVS\\
  {\tt ashim@cs.utah.edu, amrk@itu.dk,}\\{\tt
  pawang@cse.iitkgp.ac.in, oliver.hellwig@uzh.ch} \\
  }

\date{}

\begin{document}
\maketitle
\begin{abstract}
 Neural sequence labelling approaches have achieved state of the art results in morphological tagging. We evaluate the efficacy of four standard sequence labelling models on Sanskrit, a morphologically rich, fusional Indian language. As its label space can theoretically contain more than 40,000 labels, systems that explicitly model the internal structure of a label are more suited for the task, because of their ability to generalise to labels not seen during training. We find that although some neural models perform better than others, one of the common causes for error for all of these models is mispredictions due to syncretism.\footnote{Code and data available at \url{https://github.com/ashim95/sanskrit-morphological-taggers}}
\end{abstract}

\section{Introduction}

Sanskrit is a fusional Indo-European language with rich morphology, both at the inflectional and derivational level. The language relies heavily on morphological markers to determine the syntactic, and to some extent the semantic roles, of words in a sentence. There exist limited and partly incompatible solutions \cite{hellwig2016improving,goyal2016design,D18-1276krishna} for morphological tagging of Sanskrit that heavily rely on lexicon driven shallow parsers and other linguistic knowledge. However recently, neural sequential labelling models have achieved competitive results in morphological tagging for multiple languages \cite{D17-1078cottterrel, K18-1036composite, P18-1247fcrf}. We therefore explore the efficacy of such models in performing morphological tagging for Sanskrit without access to extensive linguistic information.

Most recent research treats morphological tagging as a structured prediction problem where the morphological class of a word is either treated as a monolithic label or as a composite label with multiple features \cite{muller2013efficient,D17-1078cottterrel}. \newcite{Schmid:2008:ECP:1599081.1599179,hakkani2002statistical} model the morphological tags as a sequence of individual morphological features. Recently, \newcite{K18-1036composite} proposed to generate this sequence of morphological features using a neural encoder-decoder architecture. \newcite{hellwig2016improving} shows a significant improvement in performance for morphological tagging in Sanskrit by using a monolithic tagset with recurrent neural network based tagging model. In systems using monolithic labels, multiple feature values pertaining to a word are combined to form a single label \cite{muller2013efficient,heigold2017extensive}, which leads to data sparsity for morphologically rich languages such as Czech, Turkish and Sanskrit. The sparsity issue can be addressed by using composite labels which model the internal structure of a class as a set of individual features \cite{K18-1036composite,D17-1073dontthrow}. \newcite{P18-1247fcrf} use a neural factorial CRF to model inter-dependence between individual categories of the composite morphological label. 
However, as the decision for monolithic vs. composite labels is one of the central design choices when tagging morphologically rich languages, we use Sanskrit as a test case for a systematic evaluation for this choice. For this evaluation, we consider several neural architectures with different modelling principles. For the monolithic tag model, the neural architecture is based on a bi-directional LSTM with a linear CRF layer stacked on top of it \cite{lample2016neural,huang2015bidirectional}. For composite labels, we explored a neural generation model that generates a sequence of morphological features for each word in the input sequence. In order to explicitly capture the inter-dependencies between the morphological features, we use a model based on a factorial Conditional Random Field (CRF) \cite{P18-1247fcrf}. Additionally, independent classifiers trained under a multi-task setting with sharing of parameters are also explored \cite{inoue2017joint,P16-2038}. 
Our experiments specifically focus on the following problems and questions:
%in their architectures, input representation, learning and inference. But, we find that, all the models we experimented with achieve competitive results when provided with sufficient training data. In fact, we can find uniformity in the mispredictions made by these systems. We specifically performed our experiments on the following three aspects:

\begin{itemize}[leftmargin=*]
    \item \textbf{Syncretism:} We will show that syncretism, i.e., inflected forms of a lemma that share the same surface form for multiple morphological tags, is the major source of mispredictions. We evaluate if and how models with monolithic and composite labels deal with this phenomenon.
    
    %What is the extent of misprediction due to syncretisms in the neural sequence labelling models as noted in \newcite{D18-1276krishna}? Does breaking down the tag into component categories, reduce the extent of mispredictions due to syncretism?
    
    \vspace{-0.1cm}
    
    \item \textbf{Performance on unseen tags:} For models with composite labels, it should be possible to predict morphological classes which were not seen in the training data. Our experiments show that the performance of the systems remains more or less the same irrespective of the neural architecture. 
    \end{itemize}
    This raises an important point: Models that perform marginally better in terms of evaluation metrics are supposedly not superior, since we see similar performances on special test sets targeting particular statistical phenomenon (unseen tags) and  linguistic phenomenon (syncretism).

% Introduction

\begin{table}[]
\begin{tabular}{|l|l|ccccc|}
\hline
%\multicolumn{1}{|c|}{\textbf{Feature}} & \multicolumn{1}{c|}{\textbf{values}} & \multicolumn{1}{c|}{\textbf{Noun}} & \multicolumn{1}{c|}{\textbf{Verb (Finite)}} & \multicolumn{1}{c|}{\textbf{Participles}} & \multicolumn{1}{c|}{\textbf{Compound}} & \multicolumn{1}{c|}{\textbf{Others}} \\ \hline
{\bf Feature} & {\bf\rotatebox{90}{Values}} & {\bf\rotatebox{90}{Noun}} & {\bf\rotatebox{90}{Fin. verb}} & {\bf\rotatebox{90}{Participle}} & {\bf\rotatebox{90}{Compound}} & {\bf\rotatebox{90}{Others}} \\ \hline
Tense   & 18 & & \cmark & \cmark &  & \\ 
Case   & 8 & \cmark & & \cmark & & \\ 
Number & 3 & \cmark & \cmark & \cmark & & \\ 
Gender & 3 & \cmark & & \cmark & & \\ 
Person & 3 & & \cmark & & & \\ 
Last char. & 31 & \cmark & & \cmark & \cmark & \\ 
Other & 5 & & & & & \cmark \\ \hline
Total & 71 & 2232 & 162 & 40176 & 31 & 5 \\ \hline
\end{tabular}
\caption{Grammatical features and their distribution over inflectional classes. `Others' include forms such as infinitives, absolutives, Indeclinables, etc.}
\label{tagSet}
\end{table}
\section{Problem Formulation and Models}

\paragraph{Tagset:} Sanskrit, similar to Hungarian \cite{R13-1099magyar} and Turkish \cite{W13-3704turkish}, relies on suffixes for marking inflectional information.
As Sanskrit has a rich inflectional system, the size of the tagset plays a relevant role.
%participles and other non-finite verb roles. But, previous taggers for Sanskrit did not distinguish between participles and nouns, as participles are inflected nouns which has a conjugated form from a verb-root. But, this information is important for downstream NLP tasks such as dependency parsing. 
\newcite{hellwig2016improving} uses a tagset with 86 possible labels that merges some grammatical features based on linguistic considerations. \newcite{D18-1276krishna} use an extended tagset of 270 features, by adding the feature tense, but only to finite verbs. 
As the systems tested in this paper do not use external linguistic information that could restrict the range of applicable features, we choose a tagset consisting of the features shown in Table \ref{tagSet}, which is in principle motivated by the traditional grammatical analysis of Sanskrit \cite{apte1885student}.\footnote{Exhaustive list of inflectional tags used by three prevalent tagging schemes in Sanskrit computational linguistics:  \url{https://sanskritlibrary.org/helpmorphids.html}} As the declensional type of a noun in Sanskrit is determined by the last character in the non-inflected stem of the word, we add the last character of the stem as a morphological feature in our predictions.

\begin{figure*}
  \includegraphics[width=0.9\textwidth,trim={0 8cm 0 0}]{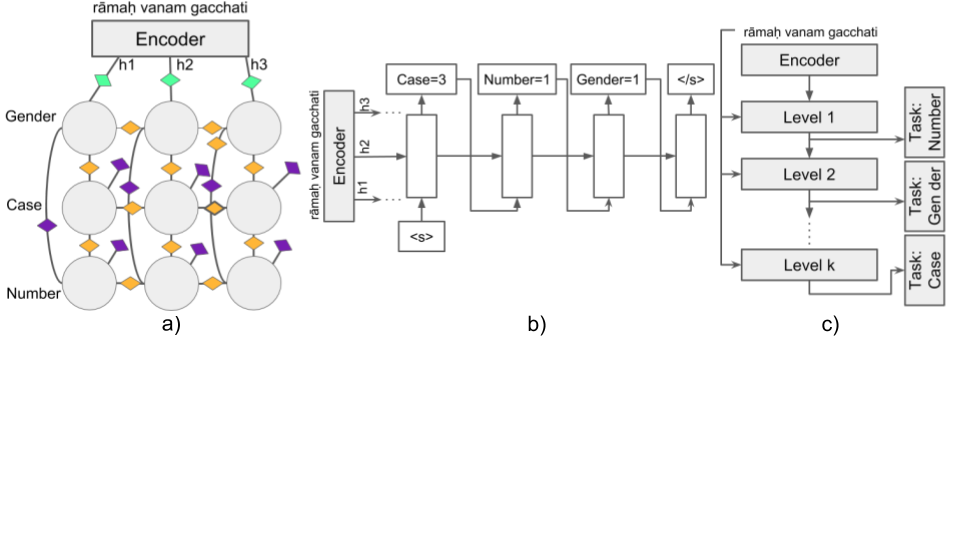}
  \caption{Models for composite tag prediction, demonstrated for the toy sentence {\sl r\=ama\d h vana\d m gacchati} (English: `R\=ama goes to the forest'). a) FCRF b) Seq c) MTL-Hierarchy. For illustrative purposes, we consider a tagset consisting of only three categories.} 
  \label{model}
\end{figure*}

%We formulate the problem as a neural sequence labelling problem, with four different architectures. We model a standard multiclass sequence labelling approach which uses the monolithic tag by combining the relevant categories as our baseline. Additionally, we experiment with the neural factor CRF approach proposed in \newcite{P18-1247fcrf}, a model that generates morphological labels as sequences as proposed in \newcite{K18-1036composite} and we also experiment with multi task setting with shared parameters as well as layer specific multitask setting as used in \newcite{P16-2038}.

%We follow the annotation scheme used in the digital corpus of Sanskrit. Table \ref{tagSet} shows the various categories that can serve as components of a morphological tag. Table \ref{pos} shows the POS tags that forms different combinations of the categories in Table \ref{tagSet}. 

%The non-fininte verbs, which consists of absolutives, infinitves, gerunds and participles are nouns derived from verbs, referred to as primary derivatives (k\d{r}danta) in Sanskrit. Hence, the last character of the lemma (of the derived noun) is important to identify the inflection of the word-form. The knwoedge of presence of verbal root is equally important in distinguishing the syntactic role the word can take in its dependency analaysis. Hence we predict the categories applied to the derived noun lemma as well as to the original verbal root. The nonfinite verbs other than participles are treated as indeclinable, and by default indeclinable does not have any additional categories to be predicted. 

\paragraph{Notation:} Given a sequence of tokens $\mathbf{x} = x_1 , x_2 ... ,x_T$, we aim at predicting a sequence of labels $ \mathbf{y} = y_1, y_2, .... y_T $, one for each token. Each label $y_i$ is a composite label $y_i = \{ y_{i0}, y_{i1}, ... y_{it} \}$ and consists of a collection of grammatical features for $x_i$.  
\paragraph{Encoder:} All neural sequence labelling models tested in this paper (see below) use an encoder that generates the input representations of words as follows.
Given a sequence of tokens as input, for each token $x_i \in \mathbf{x}$, its vector representation is obtained by concatenating its word embedding with a sub-word character embedding obtained from a bi-directional LSTM similar to \newcite{lample2016neural}. These word representations are passed through a word level Bi-LSTM to obtain hidden state $h_i$ for each token in the sequence.

%A standard LSTM based encoder architecture is used to encode the input sequence in all the models we use. For a given token, we obtain pretrained word embedding using fasttext and a character level biLSTM and then concatenate the vectors of a given token to obtain the input vector. The sequence of such input vectors are used as input to the LSTM based encoder architecture. 

\paragraph{Monolithic Sequence Model (MonSeq):} This is a standard neural sequence labelling model with a neural-CRF tagger \cite{lample2016neural,huang2015bidirectional}. A linear chain CRF is used as the output layer for the monolithic labels used. %relevant feature values are combined together to form a monolithic label for each possible value.
\paragraph{Neural Factorial CRF (FCRF):} The model proposed by \newcite{P18-1247fcrf}  is an end to end neural sequence labeling model with a factorial CRF \cite{sutton2007dynamic}.  The model is shown in Figure \ref{model}a.
% The system predicts composite labels, and the predictors for the features are linked using factors. 
In order to model the inter-dependence between different morphological types, a pairwise potential between cotemporal variables and a transitional potential between variables of the same type of tags is used. As exact inference is computationally expensive, loopy belief propagation is used to compute approximate factor and variable marginals.
% The model considers three types of potential functions for making the predictions. The pairwise links between all categories of a token capture the correlations between different features. A linear chain CRF is used as the temporal potential to capture the dependencies between tokens in the sequence. Finally the token level features are obtained from the encoder. %The relative strength between the links is learnt during training. 
% Loopy belief propagation is used for inference, as exact inference is intractable. 

%Factor graph model, where every value for each grammatical category is predicted separately. But, each grammatical category is made to be dependnet on every other categories bu forming pairwise connections between the factors. The strenth of the dependencies are learnt by the model during training.  For a given input word, tags are predicted for a subset of the factors and for the rest where not applicable NULL values are predicted. The factor connections are made such that the intra word connections span for every pair of categories and there is connection between every adjacent words for thhe same category.
\paragraph{Sequence Generation Model (Seq):} We use a sequence generation model \cite{K18-1036composite} that predicts composite labels as sequences of feature values for every token in the sequence. For token $x_i$, the LSTM hidden state $h_i$ is fed to an LSTM based decoder, which generates a sequence of feature values conditioned on the context vector $h_i$ and the previous feature value. As shown in Figure \ref{model}b, the decoding is initiated with a special marker passed as input and terminates when the decoder predicts an end marker. 
% The context vector from the encoder, $h_i$ and the label context vector from the decoder is used to predict the subsequent labels.%  The model is inspired by neural seq2seq models.

%Here the morphological tag is predicted as a sequence of category values . Similar to seq2seq models, the conditional language model uses the input context vector as the conditioning variable, and each tag is predicted where the tags predicted prior the tag is used. At decoding the prediction of a given category value depends on the previously predicted category value, previous label context vector and word's context vector.

\paragraph{Multi task learning (MTL):} Here, we consider each grammatical feature as shown in Table \ref{tagSet} as a separate task. We experiment with two settings for multi-task learning. In \textbf{MTL-Shared} the encoder parameters are shared across all the tasks, and supervision for all the tasks is performed at the same layer with each task having its own independent output CRF layer. In \textbf{MTL-Hierarchy}, as proposed by \newcite{P16-2038}, we establish a hierarchy between the grammatical categories. A hierarchical inductive bias is introduced by supervising low-level tasks at the bottom layers and higher-level tasks at the higher layers \cite{sanh2018hierarchical}. Concretely, for a task $k$, only the parameters at the output CRF layer and those at the shallower layers are updated. The model is shown in Figure \ref{model}c.
% Effectively, given a task $k$ at level $l(k)$, the parameters updated are at levels from $l(1)$ to $l(k)$ and at the CRF-inference layer for task at $l(k)$. %at the supervision activation function for the task at $l(k)$.
In order for the higher layers to have access to the inputs, we use shortcut connections as proposed in \newcite{D17-1206hashimoto}.

\section{Experiments}

\paragraph{Data:} We use a training set of 50,000 and a test set of 11,000 sentences from the Digital Corpus of Sanskrit (DCS, \newcite{hellwig_DCS}). To prepare the training set, we sample sentences such that there exist at least 100 instances for each of the 71 features in Table \ref{tagSet}. The training data still covers only 2,757 out of possible 42,606 labels, indicating that the true dimension of the target space is much lower than could be expected from Table \ref{tagSet}. The test data we use contain only about 0.5 \% of the tokens with labels not present in the training set. Additionally we use a separate \textbf{unseen} test set, where every sentence contains at least one word with a monolithic label not present in the training data.

%We prepare two sets of test data. There exists about 11 category-value pairs for which 100 instances are not available. Sentences with those category-values are kept aside as "Unseen" Test data. We created a test set which only contains those category-values, present in the training data and by default mention this as our "test data". This results in about 10,735 sentences.  

\subsection{Evaluation}
We report the performance using average token-level accuracy and F1-scores  \cite[see][]{P18-1247fcrf,D17-1078cottterrel,P16-1184botha}. The average token-level accuracy is reported on the exact match of a morphological tag for a token, i.e., if it predicts all the morphological features correctly. The F1 measure is computed on a tag-by-tag basis, i.e. macro and micro averaged at the grammatical category level, which provides partial credit to partially correct tag sets.

\subsection{Results}

\begin{table*}[]
\centering
\begin{tabular}{|c|c|c|c|c|c|c|c|c|}
\hline
\multicolumn{1}{|c|}{\textbf{System}} & \multicolumn{1}{c|}{\textbf{\begin{tabular}[c]{@{}c@{}}Token \\ Accuracy\end{tabular}}} & \multicolumn{1}{c|}{\textbf{\begin{tabular}[c]{@{}c@{}}Macro F1*\\ Score\end{tabular}}} & \multicolumn{1}{c|}{\textbf{\begin{tabular}[c]{@{}c@{}}Micro F1\\ Score\end{tabular}}} & \multicolumn{1}{c|}{\textbf{T}} & \multicolumn{1}{c|}{\textbf{C}} & \multicolumn{1}{c|}{\textbf{N}} & \multicolumn{1}{c|}{\textbf{G}} & \multicolumn{1}{c|}{\textbf{\begin{tabular}[c]{@{}c@{}}L\end{tabular}}} \\ \hline
MonSeq                                & 75.15	&85.16&	85.14	&79.64&	86.39&	91.05&	80.37 & 83.86                                                               \\ \hline
MTL-Shared                            & 81.72                                                                                 & 90.82                                                                                  & 90.84                                                                                  & 85.98                               & 88.31                              & 95.61                                & 90.84                                & 90.00                                                                                       \\ \hline
MTL-Hierarchy                         & \textbf{86.74}                                                                                 & \textbf{94.11}                                                                                  & \textbf{94.11}                                                                                  & \textbf{93.77}                               & \textbf{92.17}                      & \textbf{97.21}                                & \textbf{93.30}                                & 94.01                                                                                        \\ \hline
FCRF                                  & 85.37                                                                               & 93.01                                                                       & 93.01                                                                     & 92.35                    & 89.08                              & 96.38                                & 92.44                      & \textbf{94.03}                                                                               \\ \hline
Seq                                   & 85.95                                                                        & 92.72                                                                              & 92.72                                                                              & 91.10                                & 89.66                              & 96.52                       & 92.01                                & 93.12                                                                                        \\ \hline
\end{tabular}
\caption{Performance of different systems for the morphological tagging task. All the reported values for Macro-F1 are statistically significant $(p<0.05)$, based on pairwise t-tests between the systems. The features (refer Table \ref{tagSet}) are marked using their first letter.}
\label{mainRes}
\end{table*}

Table \ref{mainRes} shows the results for the five models studied in this paper. We find that three of the four models using composite labels obtain overall comparable results, with hierarchical Multi-task model obtaining the highest Macro and Micro F1-Scores, token accuracy as well as outperforming other models for category specific evaluation for 4 out of 5 categories.
% This model also outperforms all other models for token accuracy as well as for category specific evaluation for 4 out of 5 categories.
This highlights that there is some gains to be had by inducing a hierarchical bias among these morphological categories. As can be observed from the results, all the composite models clearly outperform MonSeq in terms of Macro- and Micro F1-Score,
\begin{table}[]
\begin{tabular}{|c|c|c|c|}
\hline
\multicolumn{1}{|l|}{} & \multicolumn{2}{c|}{Syncretism} & \multicolumn{1}{c|}{Unseen} \\ \hline
System                 & \multicolumn{2}{c|}{Macro F1}       & Macro F1                    \\ \hline
Seq                    & \multicolumn{2}{c|}{70.44}          & 55.98                       \\ \hline
MTL-Shared             & \multicolumn{2}{c|}{70.26}          & 55.02                       \\ \hline
MTL-Hierarchy          & \multicolumn{2}{c|}{70.62}          & 55.21                       \\ \hline
FCRF                   & \multicolumn{2}{c|}{68.80}          & 55.62                       \\ \hline
MonSeq                 & \multicolumn{2}{c|}{52.55}          & --                          \\ \hline
\end{tabular}
\caption{Macro-F1 score for the tokens which exhibit syncretism (top 25 pairs, based on reported mispredictions) and the unseen labels during training.}
\label{syncretism}
\end{table}
indicating their better performance for rare morphological classes. %, whereas the token accuracy of MonSeq is comparable to that of Seq.
Among the composite models, MTL-Shared clearly underperforms, which is probably due to the fact that most of its parameters are shared by all the tasks and no task specific adaptation was possible. We also perform pairwise t-tests and find that the gains reported are statistically significant ($p < 0.05$).

One of our key findings is that syncretism is a major source of error for all these systems. For the composite models, about 20 \% to 25 \% of all the mispredictions in nouns arise due to syncretism. As expected it is worse for MonSeq, where close to 37 \% mispredictions are due to this linguistic phenomenon.  
%\textbf{Further, the distribution of misprediction of a label to other labels show skewness towards labels with syncretisms???}. 
For a more detailed analysis, we check the top 25 label-pairs of mis-predictions for each system. In Table \ref{syncretism}, we report macro F1-Scores for the tokens which exhibit syncretism from this filtered set. The reported results are far below the overall F1-Macro scores, as shown in Table \ref{mainRes}.

Composite-label models are also able to make partially correct predictions for more unusual forms.
The 3rd person plural perfect form, {\sl sasarjire} (English: `they have created'), for example, is analysed as 3rd sg. perfect by Seq.
This decision should be influenced by the last letter {\sl -e}, which can indicate the 3rd singular of the perfect, while the correct, and relatively rare, affix is {\sl-ire}.
MonSeq predicts a locative singular of a non-existing noun {\sl *sasarjira} in this case. Again, this decision is probably based on the last letter {\sl -e}, which in most cases derives the locative singular.

Table \ref{syncretism} (right half) shows how the models perform for tags unseen during training (on \textbf{unseen} test set). We consider 11 of such case, number and gender combinations, 14 tense, person and number combinations and two of the tenses additionally for the participles. Among four different composite models, the 
macro F1-Score for `Seq' model is similar to what is observed in \newcite{K18-1036composite} for a fusional language like Czech. Moreover, the behaviour for all the composite models remains more or less same for unseen labels.

Next, we explore if there is a natural hierarchy for supervision of morphological categories in MTL-hierarchy model. For this, we train the system in different permutations of feature hierarchies as shown in Table \ref{mtlh}. We observe that the feature \textit{number} benefits  from supervision at a shallower level, whereas \textit{tense} always benefits from supervision at a deeper level. The trends for other features were not as conclusive, but these results show there might be an inherent hierarchy among some of the morphological features.

%In Table \ref{mtlh}, we report the performance for two features, `number' and `tense', based on training MTL-hierarchy  `Number' benefits  %The Person and Gender features were merged, as the features never co-occurred in a label. 
\begin{table}[]
\centering
\begin{tabular}{|c|c|c|}
\hline
Hierarchy & Number         & Tense          \\ \hline
N-G-C-T-L & 97.21          & \textbf{93.77} \\ \hline
T-C-N-G-L & 96.49 & 90.12          \\ \hline
N-G-L-T-C & \textbf{97.24}          & 92.85          \\ \hline
\end{tabular}
\caption{Results for Number and Tense based on different configurations in MTL-Hierarchy. The shallow to deep levels are marked by first character of the features.}
\label{mtlh}
\end{table}
\par

\newcite{krishnaThesis}, an extended version of the energy based model proposed in \newcite{D18-1276krishna}, report the state of the art results for morphological parsing in Sanskrit. They report a token level accuracy, macro averaged over sentences, of 95.33 \% on a test set of 9,576 sentences. The FCRF and SeqGen models, when tested on their test dataset, achieve an average sentence level token accuracy of 80.26 \% and 81.79 \% respectively. Here, it needs to be noted that the  morphological tagger used in \newcite{krishnaThesis} relies on a lexicon driven shallow parser to obtain a smaller search space of candidates. However, this makes the model a closed vocabulary model as it would fail for cases where the words are not recognised by the lexicon-driven parser, as there will be no analyses for such words. On the contrary, none of the models presented in this work are constrained by the vocabulary of any lexicon.

%Results reported in this paper are not directly comparable with those in state-of-the-art works \newcite{D18-1276krishna, krishnaThesis}, as the test set used by them is different. Additionally they report token level accuracy averaged over sentences while our token level accuracy is averaged over the test corpus. On their test set, our models achieve an average sentence level token accuracy between 80 \% and 82 \% (80.26 \% for FCRF, 81.79 \% for SeqGen etc.). Moreover, Owing to this dependence, their tagger is expected to  All the neural taggers studied in this paper, however, do not possess any such weakness. 
\par

\section{Conclusion and Future Work}
In this work, we evaluated various neural models for morphological tagging of Sanskrit, concentrating on models that are capable of using composite labels. We find that all the composite label models outperform MonSeq by significant margins. These models, with an exception to MTL-Shared, achieve overall competitive results when enough training data is available. A major problem for all the sequence labelling models studied in this paper is syncretism of morphological categories, which should constitute the main focus of future research.

%While the composite label models have an obvious advantage of predicting unseen labels in the training data, w, although the gains in Macro-F1 scores for the composite label models in comparison to the MonSeq model are statistically significant. 

\iffalse
\begin{figure}
  \includegraphics[width=\linewidth]{images/featerror.eps}
  \caption{Error propagation for composite labels, as expressed by token F-scores}
  \label{depsIntra}
\end{figure}
\fi

\iffalse
\begin{figure}
  \includegraphics[width=\linewidth]{images/comb_linePlot.eps}
  \caption{Influence of error position on the sentence F1-score}
  \label{deps}
\end{figure}

\begin{table}[]
\begin{tabular}{|l|ll|l|}
\hline
  & \multicolumn{2}{c|}{Mispredictions} &  Unseen \\
System & \% & Macro F1 & Macro F1 \\ \hline
Seq & 25.83 & 70.44 & 55.98 \\ 
MTL-Shared & 26.47 & 70.26 & 55.02 \\ 
MTL-Hierarchy & 21.93 & 66.60 & 53.18 \\
FCRF & 20.07 & 68.80 &  55.62 \\ 
MonSeq & 36.87 & 52.55 & -- \\\hline
\end{tabular}
\caption{Proportion of mispredictions among nouns due to syncretism (left part) and F-scores for unseen labels (right part)}
\label{syncretism}
\end{table}
\fi

\bibliographystyle{acl_natbib}
\bibliography{acl2020}

\end{document}